\documentclass[conference]{IEEEtran}
\IEEEoverridecommandlockouts
\usepackage{cite}
\usepackage{amsmath,amssymb,amsfonts}
\usepackage{algorithmic}
\usepackage{graphicx}
\usepackage{textcomp}
\usepackage{xcolor}
\usepackage{array}
\usepackage{verbatim}
\usepackage{float}
\usepackage{multirow}
\def\BibTeX{{\rm B\kern-.05em{\sc i\kern-.025em b}\kern-.08em
    T\kern-.1667em\lower.7ex\hbox{E}\kern-.125emX}}

\providecommand{\tabularnewline}{\\}
\floatstyle{ruled}
\newfloat{algorithm}{tbp}{loa}
\providecommand{\algorithmname}{Algorithm}
\floatname{algorithm}{\protect\algorithmname}

\begin{document}
\title{Estimating Bidirectional Causal Effects with Large Scale Online Kernel Learning
\thanks{This study was supported by JSPS KAKENHI Grant Number 25K21168.}
}

\author{\IEEEauthorblockN{Masahiro Tanaka}
\IEEEauthorblockA{\textit{Faculty of Economics} \\
\textit{Fukuoka University}\\
Fukuoka, Japan \\
gspddlnit45@toki.waseda.jp}
}

\maketitle

\begin{abstract}
In this study, a scalable online kernel learning framework is proposed
for estimating bidirectional causal effects in systems characterized
by mutual dependence and heteroskedasticity. Traditional causal inference
often focuses on unidirectional effects, overlooking the common bidirectional
relationships in real-world phenomena. Building on heteroskedasticity-based
identification, the proposed method integrates a quasi-maximum likelihood
estimator for simultaneous equation models with large scale online
kernel learning. It employs random Fourier feature approximations
to flexibly model nonlinear conditional means and variances, while
an adaptive online gradient descent algorithm ensures computational
efficiency for streaming and high-dimensional data. Results from extensive
simulations demonstrate that the proposed method achieves superior
accuracy and stability than single equation and polynomial approximation
baselines, exhibiting lower bias and root mean squared error across
various data-generating processes. These results confirm that the
proposed approach effectively captures complex bidirectional causal
effects with near-linear computational scaling. By combining econometric
identification with modern machine learning techniques, the proposed
framework offers a practical, scalable, and theoretically grounded
solution for large scale causal inference in natural/social science,
policy making, business, and industrial applications.
\end{abstract}


\section{Introduction}

The interest in applying big data analytics and machine learning for
causal analysis is growing steadily \cite{Guo2020,Yao2021,Nogueira2022,Weinberg2025}.
The rapid expansion and generation of large datasets present both
opportunities and challenges. While large datasets enhance the statistical
power, enabling the detection of subtle reciprocal relationships,
they require computationally efficient algorithms for handling streaming
or high-dimensional inputs without compromising interpretability.
A key methodological challenge lies in robustly extracting causal
effects from complex data while ensuring tractable estimation and
correct identification.

Understanding bidirectional causal relationships is fundamental across
natural/social science, policy making, business, and industrial applications.
Numerous real-world systems exhibit mutual dependence rather than
unidirectional causality. For example, interactions between brain
activity and behavior, predator-prey populations, policy interventions
and public responses, and employee morale and organizational performance
are mutually dependent. Despite its importance, recent research on
machine-learning-based causal inference has largely overlooked bidirectional
causal effects, focusing instead on unidirectional relationships between
variables.

To address this research gap, we propose a scalable online learning
method for bidirectional causal estimation built on heteroskedasticity-based
identification \cite{Rigobon2003}. This identification strategy can
be regarded as a variant of the instrumental variable method \cite{Angrist2001,Burgess2017,Wu2025}.
In conventional instrumental variable methods, causal parameters are
identified through exogenous shifts in the conditional mean of the
treatment variable induced by an instrument. For example, in a fish
market, the selling price and quantity are jointly determined; thus,
regressing the quantity on the price does not reveal the causal effect
of price. However, when weather conditions serve as valid instruments---correlated
with price but influencing quantity only through price changes---the
causal effect is identifiable. In essence, instrumental variable methods
exploit exogenous mean shifts in treatment variables. If an instrument
shifts the supply curve while leaving the demand curve fixed, or vice
versa, the corresponding slope can be estimated.

Heteroskedasticity-based identification, on the contrary, relies on
exogenous variations in the conditional variance of endogenous variables.
This approach estimates the entire simultaneous equation model (SEM)
in a single step, enabling the estimation of bidirectional causal
relationships. In the fish market example, an SEM comprises two equations
that describe supply and demand, respectively. If instruments influence
the variability of one equation while leaving the other unchanged,
the slopes of the corresponding curves can be identified.

Several methods for heteroskedasticity-based identification have been
proposed. The approach introduced in \cite{Rigobon2003} divides the
sample into low- and high-variance subsamples. Subsequent studies
have developed more flexible and efficient strategies, including the
control functional method \cite{Klein2010} and generalized method
of moments \cite{Lewbel2012}. This study builds on the quasi-maximum
likelihood (QML) estimator developed in \cite{Milunovich2018} because
it offers the most flexible and powerful framework and can be applied
to cases that are unidentifiable under alternative approaches. The
primary challenge in QML is specifying the conditional variance. While
domain-specific theory or the analyst's intuition may aid in modeling
the conditional mean, they provide limited guidance for modeling the
conditional variance. 

We address this challenge by integrating large scale online kernel
learning \cite{Lu2016} with the QML estimator for SEMs \cite{Milunovich2018}.
The proposed algorithm leverages kernel-based functional representations
and random Fourier feature approximations to flexibly model nonlinear
relationships in both conditional variances and means \cite{Rahimi2007}.
It combines a flexible representation with online optimization for
efficient parameter updates as new data arrive. By embedding identification
logic within a scalable learning architecture, the proposed method
bridges econometric theory and modern machine learning. The resulting
estimator captures complex bidirectional causality in the common high-dimensional
environments of contemporary empirical research. Contrary to the recent
kernel-based instrumental variable methods that estimate unidirectional
effects \cite{Singh2019,Muandet2020,Mastouri2021}, the proposed approach
jointly estimates bidirectional causal effects in a single model.

The remainder of this paper is organized as follows. Section II presents
the proposed method, including its theoretical properties and local
identification conditions. Section III details the simulation experiments
conducted to evaluate the practical performance of the proposed framework.
Finally, Section IV discusses the findings and concludes the study.

\section{Method}

\subsection{Model}

An SEM is defined as
\begin{eqnarray}
y_{1,i} & = & \gamma_{1}y_{2,i}+h_{1}\left(\boldsymbol{x}_{i},\boldsymbol{\beta}_{1}\right)+\varepsilon_{1,i},\label{eq:=000020model}\\
y_{2,i} & = & \gamma_{2}y_{1,i}+h_{2}\left(\boldsymbol{x}_{i},\boldsymbol{\beta}_{2}\right)+\varepsilon_{2,i},\nonumber 
\end{eqnarray}
for $i=1,...,n$, where $y_{1,i}$ and $y_{2,i}$ are endogenous variables,
$\boldsymbol{x}_{i}=\left(x_{1,i},....,x_{d,i}\right)^{\top}$ represents
a $d$-dimensional vector of exogenous variables, and $\varepsilon_{1,i}$
and $\varepsilon_{2,i}$ are normally distributed error terms. Functions
$h_{1}\left(\cdot\right)$ and $h_{2}\left(\cdot\right)$ are assumed
to be twice continuously differentiable, and $\gamma_{1}$, $\gamma_{2}$,
$\boldsymbol{\beta}_{1}$, and $\boldsymbol{\beta}_{2}$ are unknown
parameters. The primary objective is to estimate $\gamma_{1}$ and
$\gamma_{2}$, which capture the causal effects of $y_{2,i}$ on $y_{1,i}$
and vice versa. The conditional variances of the error terms are specified
as
\[
g_{j,i}=\mathbb{V}\left(\varepsilon_{j,i}|\boldsymbol{x}_{i}\right)=\exp\left(f_{j}\left(\boldsymbol{x}_{i},\boldsymbol{\alpha}_{j}\right)\right),\quad j=1,2,
\]
where $\boldsymbol{\alpha}_{1}$ and $\boldsymbol{\alpha}_{2}$ are
unknown parameters and every function $f_{j}\left(\cdot\right)$ is
twice continuously differentiable with respect to $\boldsymbol{\alpha}_{j}$.

Because $y_{1,i}$ and $y_{2,i}$ are introduced into the model symmetrically,
the system can be expressed as 
\begin{eqnarray*}
y_{1,i} & = & \left(-h_{2}\left(\boldsymbol{x}_{i}\right)+y_{2,i}-u_{2,i}\right)/\gamma_{2},\\
y_{2,i} & = & \left(-h_{1}\left(\boldsymbol{x}_{i}\right)+y_{1,i}+u_{1,i}\right)/\gamma_{1},
\end{eqnarray*}
for $\gamma_{1},\gamma_{2}\neq0$. The two parameterizations are observationally
equivalent, implying the existence of two possible sets of true parameter
values. Therefore, the interpretation of each equation and its parameters
depends on theoretical reasoning and prior assumptions. The following
analysis focuses on local identification. Without additional theoretical
structure, the true parameter values of the observationally equivalent
representations are treated as distinct and distant from those of
the original model.

We estimate unknown parameter vector $\boldsymbol{\theta}=\left(\gamma_{1},\gamma_{2},\boldsymbol{\beta}_{1}^{\top},\boldsymbol{\beta}_{2}^{\top},\boldsymbol{\alpha}_{1}^{\top},\boldsymbol{\alpha}_{2}^{\top}\right)^{\top}$
using a loss function derived from log-Gaussian quasi-likelihood.
Stacking the equations in (\ref{eq:=000020model}) yields
\[
\boldsymbol{\Gamma}\boldsymbol{y}_{i}=\boldsymbol{h}_{i}+\boldsymbol{\varepsilon}_{i},\quad\mathbb{V}\left[\boldsymbol{\varepsilon}_{i}|\boldsymbol{x}_{i}\right]=\boldsymbol{G}_{i}=\textrm{diag}\left(g_{1,i},g_{2,i}\right),
\]
where
\[
\boldsymbol{y}_{i}=\left(y_{1,i},y_{2,i}\right)^{\top},\quad\boldsymbol{\Gamma}=\left(\begin{array}{cc}
1 & -\gamma_{1}\\
-\gamma_{2} & 1
\end{array}\right),
\]
\[
\boldsymbol{h}_{i}=\left(h_{1}\left(\boldsymbol{x}_{i},\boldsymbol{\beta}_{1}\right),h_{2}\left(\boldsymbol{x}_{i},\boldsymbol{\beta}_{2}\right)\right)^{\top}.
\]
The log quasi-likelihood is given by
\[
\log L_{n}\left(\boldsymbol{\theta}\right)=-n\log\left(2\pi\right)+n\log\det\boldsymbol{\Gamma}
\]
\[
-\frac{1}{2}\sum_{i=1}^{n}\left[\log\det\boldsymbol{G}_{i}+\textrm{tr}\left\{ \boldsymbol{G}_{i}^{-1}\boldsymbol{\varepsilon}_{i}\left(\boldsymbol{\theta}\right)\boldsymbol{\varepsilon}_{i}\left(\boldsymbol{\theta}\right)^{\top}\right\} \right],
\]
\[
\boldsymbol{\varepsilon}_{i}\left(\boldsymbol{\theta}\right)=\left(\varepsilon_{1,i}\left(\boldsymbol{\theta}\right),\varepsilon_{2,i}\left(\boldsymbol{\theta}\right)\right)^{\top}=\boldsymbol{\Gamma}\boldsymbol{y}_{i}-\boldsymbol{h}_{i}.
\]
It can be represented as
\[
\log L_{n}\left(\boldsymbol{\theta}\right)=-n\log\left(2\pi\right)-\frac{1}{2}\sum_{i=1}^{n}\rho_{i}\left(\boldsymbol{\theta},\mathcal{D}_{i}\right),
\]
where
\begin{eqnarray*}
\rho_{i}\left(\boldsymbol{\theta},\mathcal{D}_{i}\right) & = & -2\log\left(1-\gamma_{1}\gamma_{2}\right)+\log\left(g_{1,i}g_{2,i}\right)\\
 &  & +\left(\frac{\varepsilon_{1,i}\left(\boldsymbol{\theta}\right)^{2}}{g_{1,i}}+\frac{\varepsilon_{2,i}\left(\boldsymbol{\theta}\right)^{2}}{g_{2,i}}\right),
\end{eqnarray*}
and $\mathcal{D}_{i}=\left\{ y_{1,i},y_{2,i},\boldsymbol{x}_{i}\right\} $
denotes the observations for unit $i$. Function $\rho_{i}\left(\boldsymbol{\theta},\mathcal{D}_{i}\right)$
serves as the loss function for online learning. The gradient of $\rho_{i}\left(\boldsymbol{\theta},\mathcal{D}_{i}\right)$
can be derived analytically, as shown in the Appendix.

Point identification in the proposed method relies on the following
assumptions.

\paragraph*{Assumption A}

For $i=1,....,n$, the following conditions hold:
\begin{enumerate}
\item $\det\left(\boldsymbol{\Gamma}\right)=1-\gamma_{1}\gamma_{2}\neq0$.
\item The conditional variances of the error terms are given by
\[
g_{j,i}=\mathbb{V}\left(\varepsilon_{j,i}|\boldsymbol{x}_{i}\right)=\exp\left(f_{j}\left(\boldsymbol{x}_{i}^{*},\boldsymbol{\alpha}_{j}\right)\right),\quad j=1,2,
\]
where $\boldsymbol{x}_{i}^{*}$ denotes a subvector of $\boldsymbol{x}_{i}$.
\item The conditional mean and covariance of the error terms satisfy $\mathbb{E}\left(\varepsilon_{j,i}|\boldsymbol{x}_{i}\right)=0$
for $j=1,2$ and $\mathbb{E}\left(\varepsilon_{1,i}\varepsilon_{2,i}|\boldsymbol{x}_{i}\right)=0$.
\item The standardized error terms $\varepsilon_{1,i}/\sqrt{g_{1,i}}$ and
$\varepsilon_{2,i}/\sqrt{g_{2,i}}$ are uncorrelated with $\varepsilon_{1,i^{\prime}}$
and $\varepsilon_{2,i^{\prime}}$ for $i^{\prime}\neq i$.
\item Let
\[
\nabla f_{k,i}=\frac{\partial f_{k}\left(\boldsymbol{x}_{i},\boldsymbol{\alpha}_{k}\right)}{\partial\boldsymbol{\alpha}_{k}},\quad\nabla h_{k,i}=\frac{\partial h_{k}\left(\boldsymbol{x}_{i},\boldsymbol{\beta}_{k}\right)}{\partial\boldsymbol{\beta}_{k}},
\]
\[
\mathcal{H}_{k,f}=\mathbb{E}\left[\sum_{i=1}^{n}\nabla f_{k,i}\left(\nabla f_{k,i}\right)^{\top}/n\right],
\]
and
\[
\mathcal{H}_{k,h}=\mathbb{E}\left[\sum_{i=1}^{n}\nabla h_{k,i}\left(\nabla h_{k,i}\right)^{\top}/n\right].
\]
$\mathcal{H}_{k,f}$ and $\mathcal{H}_{k,h}$ have full rank in a
neighborhood of the true parameter vector for $k=1,2$.
\end{enumerate}
According to Theorem 1 in \cite{Milunovich2018}, under Assumption
A, the true parameter vector is locally identified if and only if 
\begin{enumerate}
\item $g_{2,i}$ is not proportional to $g_{1,i}$, and
\item either 
\[
\gamma_{2}^{2}\left(1-\boldsymbol{b}_{1}^{\top}\mathcal{H}_{1,f}^{-1}\boldsymbol{b}_{1}\right)>0
\]
 or 
\[
\gamma_{1}^{2}\left(1-\boldsymbol{b}_{2}^{\top}\mathcal{H}_{2,f}^{-1}\boldsymbol{b}_{2}\right)>0,
\]
where $\boldsymbol{b}_{k}=\mathbb{E}\left[\sum_{i=1}^{n}\nabla f_{k,i}/n\right]$.
\end{enumerate}

\subsection{Specification of unknown functions}

For simplicity, we assume that unknown functions, $f_{j}\left(\cdot\right)$
and $h_{j}\left(\cdot\right)$ for $j=1,2$ depend on the same set
of covariates; that is, $\boldsymbol{x}_{i}=\boldsymbol{w}_{i}$,
thereby sharing a common learning representation of exogenous information.
We adopt a kernel-based functional approximation that maps each observation
onto feature vector $\boldsymbol{z}\left(\boldsymbol{x}\right)\in\mathbb{R}^{m}$,
induced by kernel function $\kappa\left(\cdot,\cdot\right)$ \cite{Rahimi2007}.
In this mapping, the inner product of transformed observations approximates
the kernel value as $\kappa\left(\boldsymbol{x}_{i},\boldsymbol{x}_{i^{\prime}}\right)\approx\boldsymbol{z}\left(\boldsymbol{x}_{i}\right)^{\top}\boldsymbol{z}\left(\boldsymbol{x}_{i^{\prime}}\right)$.
Using this representation, the variance function can be expressed
as
\begin{eqnarray*}
f_{j}\left(\boldsymbol{x}\right) & = & \sum_{i}\boldsymbol{\lambda}_{i}\kappa\left(\boldsymbol{x}_{i},\boldsymbol{x}\right)\\
 & \approx & \sum_{i}\boldsymbol{\lambda}_{i}\boldsymbol{z}\left(\boldsymbol{x}_{i}\right)^{\top}\boldsymbol{z}\left(\boldsymbol{x}\right)=\boldsymbol{\alpha}_{j}^{\top}\boldsymbol{z}\left(\boldsymbol{x}\right),
\end{eqnarray*}
where $\boldsymbol{\alpha}_{j}=\sum_{i}\boldsymbol{\lambda}_{i}\boldsymbol{z}\left(\boldsymbol{x}_{i}\right)$
denotes the coefficient vector in the transformed feature space. For
shift-invariant kernels, an efficient approximation is obtained through
random Fourier features. According to Bochner’s theorem, any continuous,
positive-definite, and shift-invariant kernel, $\kappa\left(\boldsymbol{x}_{1},\boldsymbol{x}_{2}\right)=\kappa\left(\boldsymbol{x}_{1}-\boldsymbol{x}_{2}\right)$,
can be expressed as the Fourier transform of a nonnegative measure:
\[
\kappa\left(\boldsymbol{x}_{1}-\boldsymbol{x}_{2}\right)=\int p\left(\boldsymbol{u}\right)\exp\left(i\boldsymbol{u}^{\top}\left(\boldsymbol{x}_{1}-\boldsymbol{x}_{2}\right)\right)d\boldsymbol{u},
\]
where $p\left(\boldsymbol{u}\right)$ is the spectral density of the
kernel obtained using the inverse Fourier transform as follows:
\[
p\left(\boldsymbol{u}\right)=\left(2\pi\right)^{-d}\int\exp\left(-i\boldsymbol{u}^{\top}\Delta\boldsymbol{x}\right)\kappa\left(\Delta\boldsymbol{x}\right)d\left(\Delta\boldsymbol{x}\right),
\]
with $\Delta\boldsymbol{x}=\boldsymbol{x}_{1}-\boldsymbol{x}_{2}$.
Rewriting the kernel as an expectation with respect to $p\left(\boldsymbol{u}\right)$,
we obtain
\[
\kappa\left(\boldsymbol{x}_{1},\boldsymbol{x}_{2}\right)=\mathbb{E}\left[\exp\left(i\boldsymbol{u}^{\top}\boldsymbol{x}_{1}\right)\exp\left(i\boldsymbol{u}^{\top}\boldsymbol{x}_{2}\right)\right].
\]
Taking its real part yields
\begin{eqnarray*}
\kappa\left(\boldsymbol{x}_{1},\boldsymbol{x}_{2}\right) & = & \mathbb{E}_{\boldsymbol{u}}\left[\cos\left(\boldsymbol{u}^{\top}\boldsymbol{x}_{1}\right)\cos\left(\boldsymbol{u}^{\top}\boldsymbol{x}_{2}\right)\right.\\
 &  & \qquad\left.+\sin\left(\boldsymbol{u}^{\top}\boldsymbol{x}_{1}\right)\sin\left(\boldsymbol{u}^{\top}\boldsymbol{x}_{2}\right)\right].
\end{eqnarray*}
Thus, the corresponding feature mapping is given by
\[
\boldsymbol{z}\left(\boldsymbol{x}\right)=\left(\sin\left(\boldsymbol{u}^{\top}\boldsymbol{x}\right),\cos\left(\boldsymbol{u}^{\top}\boldsymbol{x}\right)\right)^{\top}.
\]
To construct a finite-dimensional approximation, we independently
draw $m$ samples $\left\{ \boldsymbol{u}_{1},...,\boldsymbol{u}_{m}\right\} $
from $p\left(\boldsymbol{u}\right)$ and define
\begin{eqnarray*}
\boldsymbol{z}\left(\boldsymbol{x}\right) & = & \left(\sin\left(\boldsymbol{u}_{1}^{\top}\boldsymbol{x}\right),\cos\left(\boldsymbol{u}_{1}^{\top}\boldsymbol{x}\right),\right.\\
 &  & \quad\left....,\sin\left(\boldsymbol{u}_{m}^{\top}\boldsymbol{x}\right),\cos\left(\boldsymbol{u}_{m}^{\top}\boldsymbol{x}\right)\right)^{\top}.
\end{eqnarray*}
This random Fourier mapping efficiently approximates the kernel inner
product in a low-dimensional Euclidean space. Analogously, the conditional
mean functions are specified as $h_{j}\left(\boldsymbol{x}\right)=\boldsymbol{\beta}_{j}^{\top}\boldsymbol{z}\left(\boldsymbol{x}\right)$
for $j=1,2$.

The proposed specification corresponds to case (i) from Corollary
1 in \cite{Milunovich2018} because the conditional variance models
are defined as $g_{j,i}=\exp\left(\boldsymbol{\alpha}_{j}^{\top}\boldsymbol{z}\left(\boldsymbol{x}_{i}\right)\right)$
for $j=1,2$. Hence, the true value of $\boldsymbol{\theta}$ is locally
identified if and only if $\boldsymbol{\alpha}_{1}\neq\boldsymbol{\alpha}_{2}$.

\subsection{Computation}

We estimate parameter vector $\boldsymbol{\theta}$ using an online
gradient descent algorithm. Although several variants are available,
we select the implementation that proceeds as follows. At every iteration
$i$, gradient of loss function $\nabla\rho\left(\boldsymbol{\theta}_{i},\mathcal{D}_{i}\right)$
is scaled using adaptive gradient clipping as follows \cite{Brock2021}:
\[
\nabla\rho^{*}\left(\boldsymbol{\theta}_{i,}\right)=\nabla\rho\left(\boldsymbol{\theta}_{i,}\right)\min\left\{ 1,\;\frac{\mu_{i}}{\left\Vert \nabla\rho_{i}\left(\boldsymbol{\theta}_{i,}\right)\right\Vert _{2}}\right\} ,
\]
where $\boldsymbol{\theta}_{i}$ is the current parameter estimate,
$\mu_{i}\left(>0\right)$ is a clipping threshold, and $\left\Vert \cdot\right\Vert _{2}$
denotes the Euclidean norm. Threshold $\mu_{i}$ is updated as a bias-corrected
exponential moving average of past gradient norms:
\[
\mu_{i}=\frac{a_{i}}{1-\nu^{i}},\quad a_{i}=\nu a_{i-1}+\left(1-\nu\right)\left\Vert \nabla\rho_{i}\left(\boldsymbol{\theta}_{i}\right)\right\Vert _{2},
\]
where $\nu\in\left(0,1\right)$ is a tuning parameter. The step size
is adaptively tuned using Adam optimization \cite{Kingma2015}. As
Gaussian kernel, 
\[
\kappa\left(\boldsymbol{x}_{1},\boldsymbol{x}_{2}\right)=\exp\left(-\tau^{-1}\left\Vert \boldsymbol{x}_{1}-\boldsymbol{x}_{2}\right\Vert _{2}^{2}\right),
\]
is adopted, $p\left(\boldsymbol{u}\right)=\mathcal{N}\left(0,\tau^{-1}\boldsymbol{I}\right)$.
Kernel bandwidth $\tau\left(>0\right)$ is selected using the following
median heuristic \cite{Flaxman2016,Garreau2017}:
\[
\tau=\textrm{median}\left\{ \left\Vert \boldsymbol{x}_{i}-\boldsymbol{x}_{i^{\prime}}\right\Vert _{2}:i,i^{\prime}\in\mathcal{I}^{\dagger}\right\} ,
\]
where $\mathcal{I}^{\dagger}$ is an index set, such as an initial
batch or random subset of the full dataset.

\begin{algorithm}
\caption{Online gradient descent for estimation of bidirectional causal effects}

\medskip{}

\texttt{Input:} initial parameter value $\boldsymbol{\theta}_{init}$,
number of Fourier components $m$, and tuning parameters $\tau,\nu$.

Initialize $\boldsymbol{\theta}_{1}=\boldsymbol{\theta}_{\textrm{init}}$,
$a_{0}=0$.

Sample $\left\{ \boldsymbol{u}_{1},...,\boldsymbol{u}_{m}\right\} $
from $p\left(\boldsymbol{u}\right)$.

\texttt{for} $i=1,2,...,N$\texttt{:}

$\quad$Construct feature representation as

$\quad$$\quad$$\boldsymbol{z}\left(\boldsymbol{x}\right)=\left(\sin\left(\boldsymbol{u}_{1}^{\top}\boldsymbol{x}\right),\cos\left(\boldsymbol{u}_{1}^{\top}\boldsymbol{x}\right),\right.$

$\quad$$\quad$$\quad$$\quad$$\quad$$\quad$$\left....,\sin\left(\boldsymbol{u}_{m}^{\top}\boldsymbol{x}\right),\cos\left(\boldsymbol{u}_{m}^{\top}\boldsymbol{x}\right)\right)^{\top}.$

$\quad$Update step size $\eta_{i}$ using Adam optimization \cite{Kingma2015}.

$\quad$Update moving average as follows:

$\quad$$\quad$$\mu_{i}=\frac{a_{i}}{1-\nu^{i}}$, $a_{i}=\nu a_{i-1}+\left(1-\nu\right)\left\Vert \nabla\rho_{i}\left(\boldsymbol{\theta}_{i}\right)\right\Vert _{2}$.

$\quad$Compute clipped gradient as follows:

$\quad$$\quad$$\nabla\rho_{i}^{*}\left(\boldsymbol{\theta}_{i}\right)=\nabla\rho_{i}\left(\boldsymbol{\theta}_{i}\right)\min\left\{ 1,\;\frac{\mu_{i}}{\left\Vert \nabla\rho_{i}\left(\boldsymbol{\theta}_{i}\right)\right\Vert _{2}}\right\} $.

$\quad$Update parameters using $\boldsymbol{\theta}_{i+1}=\boldsymbol{\theta}_{i}-\eta_{i}\nabla\rho_{i}^{*}\left(\boldsymbol{\theta}_{i}\right)$.

\texttt{end for}
\end{algorithm}

\section{Experiment}

To evaluate the proposed method, we conducted a simulation study comparing
three alternative methods. 
\begin{enumerate}
\item \textbf{SEM-Kernel}: Proposed method. 
\item \textbf{Single-Kernel}: Models the mean effect using the same kernel
approximation as SEM-Kernel but estimates each equation independently
via the following quadratic loss function:
\[
\rho\left(\boldsymbol{\theta}_{1,i},\mathcal{D}_{i}\right)=\left(y_{1,i}-\gamma_{1}y_{2,i}-h_{1}\left(\boldsymbol{x}_{i},\boldsymbol{\beta}_{1}\right)\right)^{2},
\]
with an analogous specification for the second equation. 
\item \textbf{SEM-PAB}: Employs a polynomial approximation with beta function
weights and a Box--Cox transformation, corresponding to the most
flexible specification in \cite{Milunovich2018}. The conditional
variance models are defined as
\[
g_{j,i}=\begin{cases}
\frac{\left(\exp\left(g_{j,i}^{*}\right)\right)^{\breve{\alpha}_{j}}-1}{\breve{\alpha}_{j}}, & \breve{\alpha}_{j}\neq0,\\
g_{j,i}^{*}, & \breve{\alpha}_{j}=0,
\end{cases}
\]
\[
g_{j,i}^{*}=\sum_{l=1}^{d}\tilde{g}_{j,i,l}^{2},
\]
\begin{eqnarray*}
\tilde{g}_{j,i,l} & = & \exp\left(\alpha_{j,l,0}\right)\\
 &  & +\exp\left(\alpha_{j,l,1}\right)\sum_{r=1}^{4}b_{j,l,r}\left(\alpha_{j,l,2},\alpha_{j,l,3}\right)x_{l,i},
\end{eqnarray*}
\[
b_{j,l,r}\left(\alpha_{j,l,2},\alpha_{j,l,3}\right)=
\]
\[
\frac{\left(\frac{r}{4+1}\right)^{\alpha_{j,l,2}-1}\left(1-\frac{r}{4+1}\right)^{\alpha_{j,l,3}-1}}{\sum_{r^{\prime}=1}^{4}\left(\left(\frac{r^{\prime}}{4+1}\right)^{\alpha_{j,l,2}-1}\left(1-\frac{r^{\prime}}{4+1}\right)^{\alpha_{j,l,3}-1}\right)},
\]
for $l=1,....,d$ and $j=1,2$. Thus, $\boldsymbol{\alpha}_{j}=\left(\boldsymbol{\alpha}_{j,1}^{\top},...,\boldsymbol{\alpha}_{j,d}^{\top},\breve{\alpha}_{j}\right)^{\top}$,
with 
\[
\boldsymbol{\alpha}_{j,l}=\left(\alpha_{j,l,0},\alpha_{j,l,1},\alpha_{j,l,2},\alpha_{j,l.3}\right)^{\top}.
\]
To ensure the positivity of $g_{j,i}$, parameters ($\alpha_{j,l,0},\alpha_{j,l,1}$)
are introduced into the model through exponentiation. The conditional
mean functions are defined similarly but with a simpler formulation
because they are unconstrained:
\[
h_{j,i}=\sum_{l=1}^{d}\beta_{j,l,0}+\beta_{j,l,1}\sum_{r=1}^{4}b_{j,l,r}\left(\beta_{j,l,2},\beta_{j,l,3}\right)x_{l,i},
\]
where 
\[
b_{j,l,r}\left(\beta_{j,l,2},\beta_{j,l,3}\right)=
\]
\[
\frac{\left(\frac{r}{4+1}\right)^{\beta_{j,l,2}-1}\left(1-\frac{r}{4+1}\right)^{\beta_{j,l,3}-1}}{\sum_{r^{\prime}=1}^{4}\left(\left(\frac{r^{\prime}}{4+1}\right)^{\beta_{j,l,2}-1}\left(1-\frac{r^{\prime}}{4+1}\right)^{\beta_{j,l,3}-1}\right)}
\]
and $\boldsymbol{\beta}_{j}=\left(\boldsymbol{\beta}_{j,1}^{\top},...,\boldsymbol{\beta}_{j,d}^{\top}\right)^{\top}$,
with 
\[
\boldsymbol{\beta}_{j,l}=\left(\beta_{j,l,0},\beta_{j,l,1},\beta_{j,l,2},\beta_{j,l.3}\right)^{\top}.
\]
\end{enumerate}
Synthetic data were generated according to (\ref{eq:=000020model}).
The true causal parameters were fixed to $\gamma_{1}=-0.5$ and $\gamma_{2}=1.0$,
as in \cite{Milunovich2018}. The exogenous variables were drawn from
a zero-mean multivariate normal distribution, $\boldsymbol{x}_{i}\sim\mathcal{N}\left(\boldsymbol{0}_{d},\boldsymbol{S}\right)$.
Correlation matrix $\boldsymbol{S}$ was randomly generated from an
inverse Wishart distribution with identity scaling and $d+1$ degrees
of freedom, $\boldsymbol{S}\sim\mathcal{IW}\left(\boldsymbol{I}_{d},d+1\right)$.
The resulting matrix was normalized as $\boldsymbol{S}\leftarrow\bar{\boldsymbol{S}}\boldsymbol{S}\bar{\boldsymbol{S}}$,
where $\bar{\boldsymbol{S}}=\textrm{diag}\left(s_{1,1}^{-1/2},...,s_{d,d}^{-1/2}\right)$.
We set $d=100$ and examined three data-generating processes (DGPs).
DGP-1 and DGP-2 follow the specifications in \cite{Milunovich2018},
while DGP-3 employs more complex functional forms inspired by \cite{Chib2013}.

DGP-1:
\[
h_{1,i}=0.5+0.8x_{1,i},\quad h_{2,i}=0.5+0.8x_{1,i},
\]
\[
g_{1,i}=0.1+0.9x_{1,i}^{2},\quad g_{2,i}=0.3+0.5x_{1,i}^{2}.
\]

DGP-2:
\[
h_{1,i}=0.5+0.8x_{1,i},\quad h_{2,i}=0.5+0.8x_{1,i},
\]
\[
g_{1,i}=\exp\left(0.1+0.9x_{1,i}\right),\quad g_{2,i}=\exp\left(0.3+0.5x_{1,i}\right).
\]

DGP-3:
\[
h_{1,i}=x_{1,i}+2\exp\left(-16x_{1,i}^{2}\right)+1.5x_{2,i},
\]
\begin{eqnarray*}
h_{2,i} & = & \frac{1}{2}\left(\phi\left(x_{1,i};0.2,0.04\right)+\phi\left(x_{1,i};0.6,0.1\right)\right)\\
 &  & +1+\sin\left(2\pi x_{2,i}\right),
\end{eqnarray*}
\[
g_{1,i}=\exp\left(\log\left(0.5\right)-\frac{1}{8}x_{1,i}^{2}+x_{2,i}+\sin\left(4\pi x_{2,i}\right)\right),
\]
\[
g_{2,i}=\exp\left(-2.7-x_{1,i}+\exp\left(-50\left(x_{1,i}-0.5\right)^{2}\right)+x_{2,i}\right),
\]
where $\phi\left(x;a,b\right)$ denotes the probability density function
of a normal distribution with mean $a$ and variance $b$ evaluated
at $x$. Two independent chi-squared random variables with 10 degrees
of freedom, $\tilde{\varepsilon}_{j,1},...,\tilde{\varepsilon}_{j,n}$,
were generated and normalized to have zero mean and unit variance,
$\tilde{\varepsilon}_{j,i}\leftarrow\left(\tilde{\varepsilon}_{j,i}-10\right)/\sqrt{20}$.
Structural errors were computed as $\varepsilon_{j,i}=\tilde{\varepsilon}_{j,i}\sqrt{g_{j,i}}$.
The number of observations and features were set to $n\in\left\{ 5000,\;20,000\right\} $
and $d\in\left\{ 100,\;1000\right\} $, respectively. We set $\nu=0.99$
and used the Adam optimization hyperparameters from the original study
\cite{Kingma2015}. The model performance was evaluated in terms of
the mean bias, standard deviation (s.d.), and root mean squared error
(RMSE) of parameter estimates across 1000 Monte Carlo replications.

Tables I and II list the results for $n=5000$ and $n=20,000$, respectively,
with $m=500$. Across the three DGPs and sample sizes, the proposed
SEM-Kernel method consistently outperformed both baselines in terms
of bias and RMSE. The improvement was most pronounced for $d=100$,
demonstrating the scalability and robustness of the kernel representation
in high-dimensional settings. The Single-Kernel method, which ignored
the simultaneous equation structure, exhibited systematic bias, confirming
that neglecting the endogeneity between $y_{1}$ and $y_{2}$ leads
to inconsistent estimates, even under flexible nonparametric specifications.
The SEM-PAB method was theoretically capable of modeling complex nonlinearities,
but showed numerical instability. Overall, these results indicate
that the proposed online kernel learning method achieves lower estimation
errors and more stable convergence than the comparison methods across
Monte Carlo replications. The performance gains were particularly
strong under complex heteroskedastic structures (DGP-3), suggesting
that random feature approximation captures local smoothness and heterogeneity
in conditional variances.

\begin{table}
\caption{Simulation results (1) for $n=5000$}

\medskip{}

\centering{}%
\begin{tabular}{lrlr@{\extracolsep{0pt}.}lr@{\extracolsep{0pt}.}lr@{\extracolsep{0pt}.}lr@{\extracolsep{0pt}.}l}
\hline 
DGP & $d$ & Method  & \multicolumn{4}{c}{$\gamma_{1}$} & \multicolumn{4}{c}{$\gamma_{2}$}\tabularnewline
\cline{4-11}
 &  &  & \multicolumn{2}{c}{Bias} & \multicolumn{2}{c}{RMSE} & \multicolumn{2}{c}{Bias} & \multicolumn{2}{c}{RMSE}\tabularnewline
 &  &  & (s&d.) & \multicolumn{2}{c}{} & (s&d.) & \multicolumn{2}{c}{}\tabularnewline
\hline 
\multirow{12}{*}{DGP-1} & \multirow{6}{*}{100} & SEM-Kernel & -0&003 & 0&178 & -0&011 & 0&180\tabularnewline
 &  &  & (0&178) & \multicolumn{2}{c}{} & (0&180) & \multicolumn{2}{c}{}\tabularnewline
 &  & Single-Kernel & 0&254 & 0&295 & -0&164 & 0&210\tabularnewline
 &  &  & (0&150) & \multicolumn{2}{c}{} & (0&132) & \multicolumn{2}{c}{}\tabularnewline
 &  & SEM-PAB & 1&212 & 1&236 & 0&333 & 0&474\tabularnewline
 &  &  & (0&242) & \multicolumn{2}{c}{} & (0&337) & \multicolumn{2}{c}{}\tabularnewline
\cline{2-11}
 & \multirow{6}{*}{1000} & SEM-Kernel & 0&002 & 0&178 & -0&002 & 0&181\tabularnewline
 &  &  & (0&178) & \multicolumn{2}{c}{} & (0&181) & \multicolumn{2}{c}{}\tabularnewline
 &  & Single-Kernel & 0&251 & 0&291 & -0&161 & 0&208\tabularnewline
 &  &  & (0&148) & \multicolumn{2}{c}{} & (0&132) & \multicolumn{2}{c}{}\tabularnewline
 &  & SEM-PAB & 1&212 & 1&229 & 0&359 & 0&465\tabularnewline
 &  &  & (0&205) & \multicolumn{2}{c}{} & (0&295) & \multicolumn{2}{c}{}\tabularnewline
\hline 
\multirow{12}{*}{DGP-2} & \multirow{6}{*}{100} & SEM-Kernel & -0&004 & 0&178 & -0&009 & 0&176\tabularnewline
 &  &  & (0&178) & \multicolumn{2}{c}{} & (0&176) & \multicolumn{2}{c}{}\tabularnewline
 &  & Single-Kernel & 0&327 & 0&351 & -0&248 & 0&273\tabularnewline
 &  &  & (0&127) & \multicolumn{2}{c}{} & (0&115) & \multicolumn{2}{c}{}\tabularnewline
 &  & SEM-PAB & 1&213 & 1&236 & 0&333 & 0&474\tabularnewline
 &  &  & (0&242) & \multicolumn{2}{c}{} & (0&337) & \multicolumn{2}{c}{}\tabularnewline
\cline{2-11}
 & \multirow{6}{*}{1000} & SEM-Kernel & 0&002 & 0&178 & -0&004 & 0&184\tabularnewline
 &  &  & (0&178) & \multicolumn{2}{c}{} & (0&184) & \multicolumn{2}{c}{}\tabularnewline
 &  & Single-Kernel & 0&322 & 0&346 & -0&244 & 0&269\tabularnewline
 &  &  & (0&127) & \multicolumn{2}{c}{} & (0&114) & \multicolumn{2}{c}{}\tabularnewline
 &  & SEM-PAB & 1&212 & 1&230 & 0&359 & 0&464\tabularnewline
 &  &  & (0&205) & \multicolumn{2}{c}{} & (0&294) & \multicolumn{2}{c}{}\tabularnewline
\hline 
\multirow{12}{*}{DGP-3} & \multirow{6}{*}{100} & SEM-Kernel & -0&002 & 0&179 & -0&009 & 0&175\tabularnewline
 &  &  & (0&179) & \multicolumn{2}{c}{} & (0&175) & \multicolumn{2}{c}{}\tabularnewline
 &  & Single-Kernel & 0&300 & 0&335 & -0&100 & 0&190\tabularnewline
 &  &  & (0&149) & \multicolumn{2}{c}{} & (0&162) & \multicolumn{2}{c}{}\tabularnewline
 &  & SEM-PAB & 1&212 & 1&236 & 0&335 & 0&475\tabularnewline
 &  &  & (0&242) & \multicolumn{2}{c}{} & (0&337) & \multicolumn{2}{c}{}\tabularnewline
\cline{2-11}
 & \multirow{6}{*}{1000} & SEM-Kernel & 0&003 & 0&177 & -0&002 & 0&183\tabularnewline
 &  &  & (0&177) & \multicolumn{2}{c}{} & (0&184) & \multicolumn{2}{c}{}\tabularnewline
 &  & Single-Kernel & 0&303 & 0&339 & -0&097 & 0&195\tabularnewline
 &  &  & (0&152) & \multicolumn{2}{c}{} & (0&170) & \multicolumn{2}{c}{}\tabularnewline
 &  & SEM-PAB & 1&214 & 1&231 & 0&360 & 0&465\tabularnewline
 &  &  & (0&204) & \multicolumn{2}{c}{} & (0&294) & \multicolumn{2}{c}{}\tabularnewline
\hline 
\end{tabular}
\end{table}

\begin{table}
\caption{Simulation results (2) for $n=20,000$}

\medskip{}

\centering{}%
\begin{tabular}{lrlr@{\extracolsep{0pt}.}lr@{\extracolsep{0pt}.}lr@{\extracolsep{0pt}.}lr@{\extracolsep{0pt}.}l}
\hline 
DGP & $d$ & Method & \multicolumn{4}{c}{$\gamma_{1}$} & \multicolumn{4}{c}{$\gamma_{2}$}\tabularnewline
\cline{4-11}
 &  &  & \multicolumn{2}{c}{Bias} & \multicolumn{2}{c}{RMSE} & \multicolumn{2}{c}{Bias} & \multicolumn{2}{c}{RMSE}\tabularnewline
 &  &  & (s&d.) & \multicolumn{2}{c}{} & (s&d.) & \multicolumn{2}{c}{}\tabularnewline
\hline 
\multirow{12}{*}{DGP-1} & \multirow{6}{*}{100} & SEM-Kernel & -0&007 & 0&179 & -0&026 & 0&301\tabularnewline
 &  &  & (0&179) & \multicolumn{2}{c}{} & (0&300) & \multicolumn{2}{c}{}\tabularnewline
 &  & Single-Kernel & 0&732 & 0&734 & -0&460 & 0&463\tabularnewline
 &  &  & (0&062) & \multicolumn{2}{c}{} & (0&052) & \multicolumn{2}{c}{}\tabularnewline
 &  & SEM-PAB & 1&215 & 1&236 & 0&339 & 0&467\tabularnewline
 &  &  & (0&228) & \multicolumn{2}{c}{} & (0&322) & \multicolumn{2}{c}{}\tabularnewline
\cline{2-11}
 & \multirow{6}{*}{1000} & SEM-Kernel & 0&007 & 0&180 & -0&047 & 0&311\tabularnewline
 &  &  & (0&180) & \multicolumn{2}{c}{} & (0&307) & \multicolumn{2}{c}{}\tabularnewline
 &  & Single-Kernel & 0&721 & 0&724 & -0&449 & 0&453\tabularnewline
 &  &  & (0&063) & \multicolumn{2}{c}{} & (0&057) & \multicolumn{2}{c}{}\tabularnewline
 &  & SEM-PAB & 1&214 & 1&236 & 0&335 & 0&469\tabularnewline
 &  &  & (0&235) & \multicolumn{2}{c}{} & (0&329) & \multicolumn{2}{c}{}\tabularnewline
\hline 
\multirow{12}{*}{DGP-2} & \multirow{6}{*}{100} & SEM-Kernel & -0&007 & 0&180 & -0&021 & 0&269\tabularnewline
 &  &  & (0&180) & \multicolumn{2}{c}{} & (0&269) & \multicolumn{2}{c}{}\tabularnewline
 &  & Single-Kernel & 0&747 & 0&748 & -0&548 & 0&550\tabularnewline
 &  &  & (0&039) & \multicolumn{2}{c}{} & (0&043) & \multicolumn{2}{c}{}\tabularnewline
 &  & SEM-PAB & 1&215 & 1&236 & 0&338 & 0&467\tabularnewline
 &  &  & (0&228) & \multicolumn{2}{c}{} & (0&321) & \multicolumn{2}{c}{}\tabularnewline
\cline{2-11}
 & \multirow{6}{*}{1000} & SEM-Kernel & 0&007 & 0&180 & -0&034 & 0&248\tabularnewline
 &  &  & (0&180) & \multicolumn{2}{c}{} & (0&246) & \multicolumn{2}{c}{}\tabularnewline
 &  & Single-Kernel & 0&741 & 0&742 & -0&542 & 0&544\tabularnewline
 &  &  & (0&041) & \multicolumn{2}{c}{} & (0&046) & \multicolumn{2}{c}{}\tabularnewline
 &  & SEM-PAB & 1&214 & 1&236 & 0&334 & 0&470\tabularnewline
 &  &  & (0&235) & \multicolumn{2}{c}{} & (0&331) & \multicolumn{2}{c}{}\tabularnewline
\hline 
\multirow{12}{*}{DGP-3} & \multirow{6}{*}{100} & SEM-Kernel & -0&007 & 0&181 & -0&011 & 0&209\tabularnewline
 &  &  & (0&181) & \multicolumn{2}{c}{} & (0&209) & \multicolumn{2}{c}{}\tabularnewline
 &  & Single-Kernel & 0&784 & 0&787 & -0&312 & 0&332\tabularnewline
 &  &  & (0&067) & \multicolumn{2}{c}{} & (0&112) & \multicolumn{2}{c}{}\tabularnewline
 &  & SEM-PAB & 1&215 & 1&236 & 0&340 & 0&470\tabularnewline
 &  &  & (0&228) & \multicolumn{2}{c}{} & (0&324) & \multicolumn{2}{c}{}\tabularnewline
\cline{2-11}
 & \multirow{6}{*}{1000} & SEM-Kernel & 0&008 & 0&179 & -0&034 & 0&250\tabularnewline
 &  &  & (0&179) & \multicolumn{2}{c}{} & (0&248) & \multicolumn{2}{c}{}\tabularnewline
 &  & Single-Kernel & 0&781 & 0&784 & -0&299 & 0&320\tabularnewline
 &  &  & (0&073) & \multicolumn{2}{c}{} & (0&113) & \multicolumn{2}{c}{}\tabularnewline
 &  & SEM-PAB & 1&214 & 1&236 & 0&334 & 0&469\tabularnewline
 &  &  & (0&235) & \multicolumn{2}{c}{} & (0&330) & \multicolumn{2}{c}{}\tabularnewline
\hline 
\end{tabular}
\end{table}

We conducted a sensitivity analysis on the number of random Fourier
components, $m\in\left\{ 100,\;200,\;500,\;1000,\;2000,\;5000\right\} $,
using DGP-3. Figures 1 and 2 show the corresponding results. As expected,
both the bias and RMSE decreased rapidly as $m$ increased to approximately
500, and no notable improvement was achieved afterward. Hence, a relatively
small number of Fourier bases provides an accurate approximation of
the underlying kernel and properly balances accuracy and computational
cost. For a very large $m$, the performance gain was negligible,
while the computation time increased approximately linearly with $m$.
These findings suggest that a moderate feature dimension (e.g., $m=500-1000$)
is adequate for large scale online kernel learning. The stability
of performance across sample sizes further demonstrates that the proposed
method adapts well to streaming data without requiring recalibration
of $m$. Overall, the proposed method exhibits strong robustness to
the choice of kernel-feature dimensionality, reinforcing its practicality
for real-time causal inference in high-dimensional settings.

\begin{figure}
\caption{Sensitivity to $m$ (1) for $n=5000$}

\medskip{}

\centering{}\includegraphics[scale=0.44]{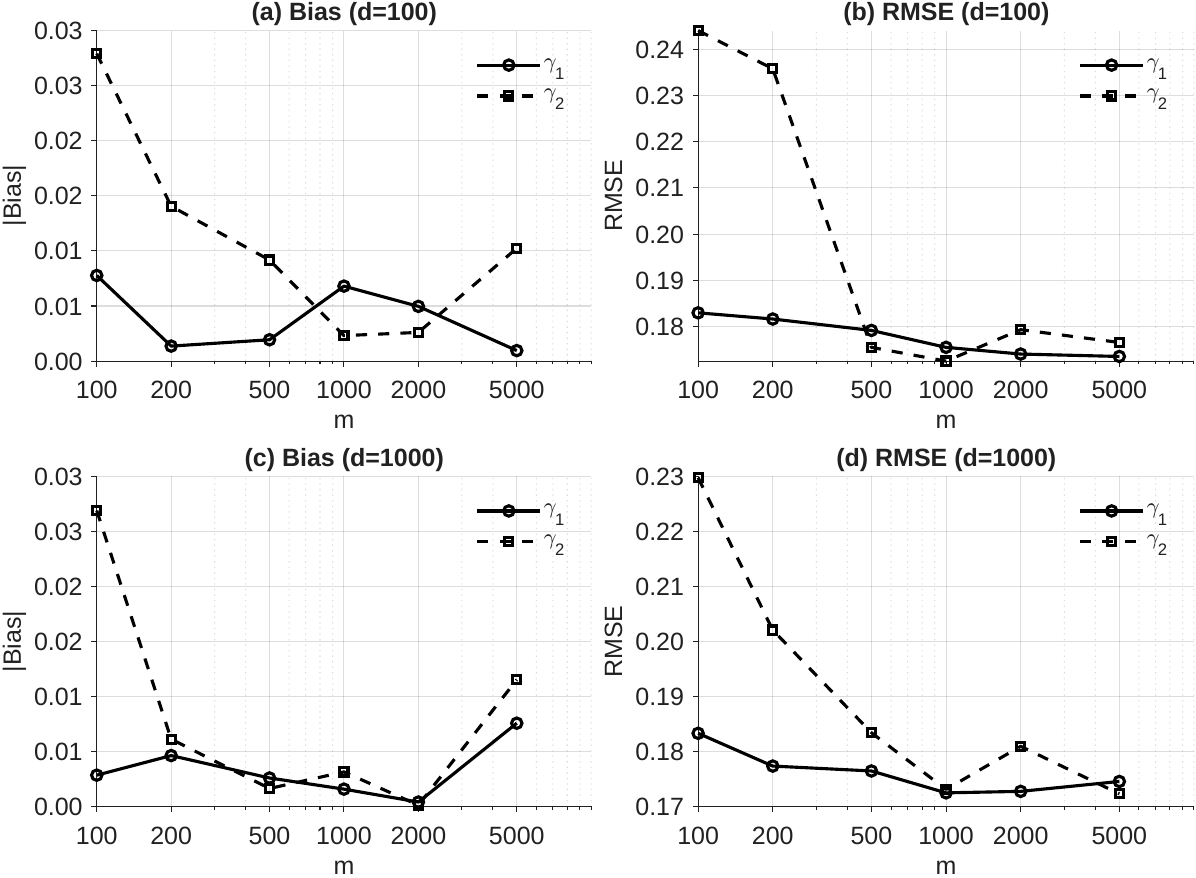}
\end{figure}

\begin{figure}
\caption{Sensitivity to $m$ (2) for $n=20,000$}

\medskip{}

\centering{}\includegraphics[scale=0.44]{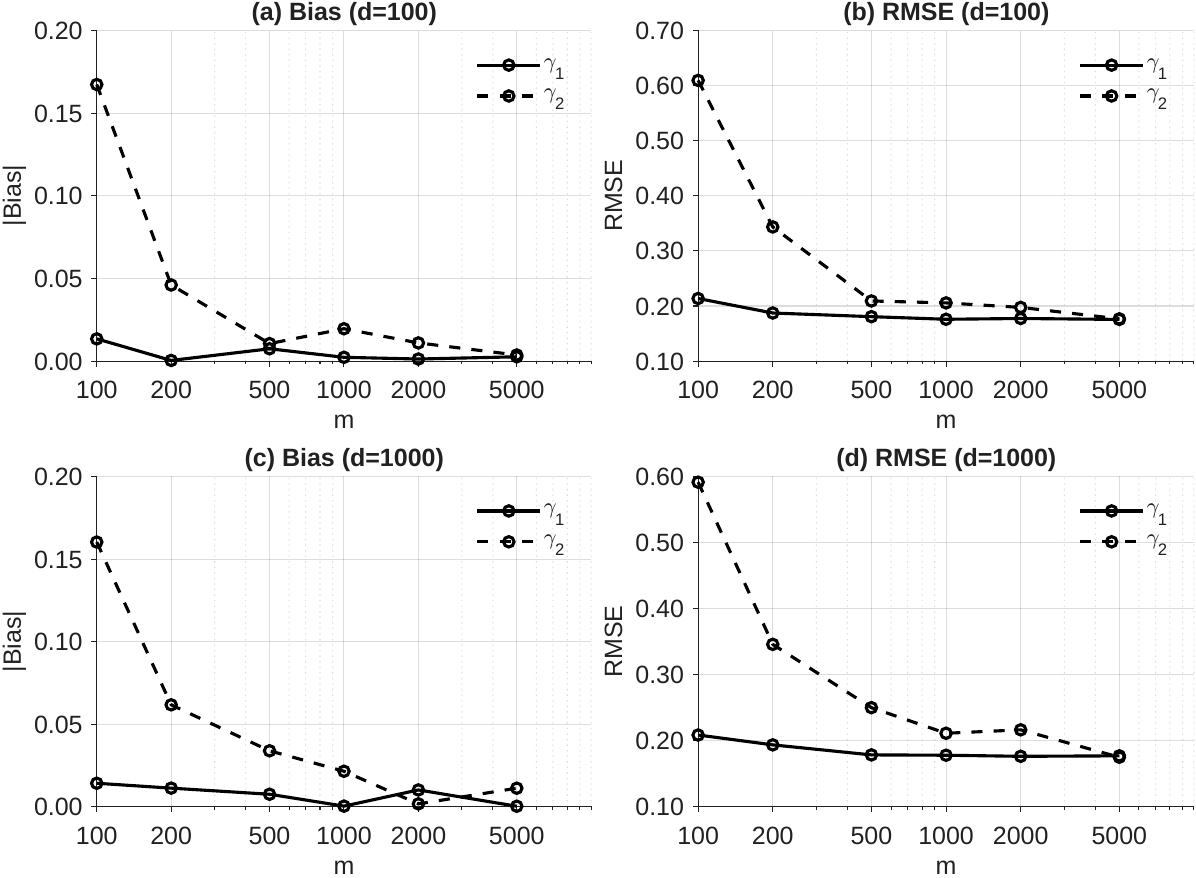}
\end{figure}

Table III lists the computation times in seconds. Despite jointly
estimating both structural equations and modeling heteroskedasticity,
the proposed SEM-Kernel method required only a slightly longer computation
time than Single-Kernel while achieving a substantially higher accuracy.
This near-parity in computational speed arises from the use of random
Fourier features and online gradient descent, which scale linearly
with the number of observations and covariates.

In contrast, the SEM-PAB method was more than an order of magnitude
slower, particularly for $d=1000$, reflecting the high computational
burden of high-dimensional polynomial expansions and Box--Cox transformations.
The computation time of SEM-Kernel increased only modestly with the
sample size, from approximately 0.7 s for $n=5000$ to 4.5 s for $n=20,000$,
thereby confirming its scalability for large streaming datasets. Overall,
the evaluation results demonstrate that the proposed method properly
balances statistical precision and computational efficiency, making
it suitable for large scale, high-dimensional causal inference.

\begin{table}

\caption{Computation time}

\medskip{}

\centering{}%
\begin{tabular}{lrlr@{\extracolsep{0pt}.}lr@{\extracolsep{0pt}.}l}
\hline 
DGP & $d$ & Method & \multicolumn{4}{c}{Computation time (s)}\tabularnewline
 &  &  & \multicolumn{2}{c}{$n=5000$} & \multicolumn{2}{c}{$n=20000$}\tabularnewline
\hline 
\multirow{6}{*}{DGP-1} & \multirow{3}{*}{100} & SEM-Kernel & 0&69 & 3&39\tabularnewline
 &  & Single-Kernel & 0&45 & 2&05\tabularnewline
 &  & SEM-PAB & 2&55 & 10&45\tabularnewline
\cline{2-7}
 & \multirow{3}{*}{1000} & SEM-Kernel & 1&32 & 4&53\tabularnewline
 &  & Single-Kernel & 1&11 & 3&03\tabularnewline
 &  & SEM-PAB & 27&89 & 112&15\tabularnewline
\hline 
\multirow{6}{*}{DGP-2} & \multirow{3}{*}{100} & SEM-Kernel & 0&66 & 3&34\tabularnewline
 &  & Single-Kernel & 0&47 & 2&03\tabularnewline
 &  & SEM-PAB & 2&56 & 10&38\tabularnewline
\cline{2-7}
 & \multirow{3}{*}{1000} & SEM-Kernel & 1&33 & 4&49\tabularnewline
 &  & Single-Kernel & 1&11 & 3&01\tabularnewline
 &  & SEM-PAB & 27&96 & 112&11\tabularnewline
\hline 
\multirow{6}{*}{DGP-3} & \multirow{3}{*}{100} & SEM-Kernel & 0&66 & 3&32\tabularnewline
 &  & Single-Kernel & 0&46 & 1&99\tabularnewline
 &  & SEM-PAB & 2&57 & 10&28\tabularnewline
\cline{2-7}
 & \multirow{3}{*}{1000} & SEM-Kernel & 1&33 & 4&48\tabularnewline
 &  & Single-Kernel & 1&10 & 3&01\tabularnewline
 &  & SEM-PAB & 27&98 & 112&54\tabularnewline
\hline 
\end{tabular}
\end{table}

\section{Discussion}

A scalable online kernel learning method is proposed for estimating
bidirectional causal effects under heteroskedasticity-based identification.
By combining the random Fourier features with online optimization,
the method flexibly models nonlinear conditional structures while
maintaining computational efficiency. Simulation results demonstrate
that it consistently outperforms existing alternatives in terms of
both estimation accuracy and scalability. This method offers a practical
and theoretically grounded solution for large scale causal inference
in systems characterized by mutual dependence. By bridging econometric
identification techniques with modern machine learning methods, it
reliably estimates bidirectional causal effects in complex, high-dimensional
environments. However, the proposed method is limited by its estimation
of only linear causal effects and assumption of the symmetry of the
error terms. Therefore, extending the framework to accommodate nonlinear
causal relationships and strongly skewed data constitutes an important
direction for future research.

\appendix{}

The gradient of the loss function is computed as follows:

\[
\nabla_{\boldsymbol{\theta}}\rho_{i}\left(\boldsymbol{\theta},\mathcal{D}_{i}\right)=\left(\begin{array}{c}
\nabla_{\boldsymbol{\gamma}}\rho_{i}\left(\boldsymbol{\theta},\mathcal{D}_{i}\right)\\
\nabla_{\boldsymbol{\beta}}\rho_{i}\left(\boldsymbol{\theta},\mathcal{D}_{i}\right)\\
\nabla_{\boldsymbol{\alpha}}\rho_{i}\left(\boldsymbol{\theta},\mathcal{D}_{i}\right)
\end{array}\right),
\]
\[
\boldsymbol{\gamma}=\left(\gamma_{1,}\gamma_{2}\right)^{\top},\quad\boldsymbol{\beta}=\left(\boldsymbol{\beta}_{1}^{\top},\boldsymbol{\beta}_{2}^{\top}\right)^{\top},\quad\boldsymbol{\alpha}=\left(\boldsymbol{\alpha}_{1}^{\top},\boldsymbol{\alpha}_{2}^{\top}\right)^{\top},
\]
\[
\nabla_{\boldsymbol{\gamma}}\rho_{i}\left(\boldsymbol{\theta},\mathcal{D}_{i}\right)=\boldsymbol{R}^{\top}\textrm{vec}\left(\boldsymbol{y}_{i}\boldsymbol{\varepsilon}_{i}\left(\boldsymbol{\theta}\right)^{\top}\boldsymbol{G}_{i}^{-1}-n\boldsymbol{\Gamma}^{-1}\right),
\]
\[
\nabla_{\boldsymbol{\beta}}\rho_{i}\left(\boldsymbol{\theta},\mathcal{D}_{i}\right)=\textrm{vec}\left(\boldsymbol{z}_{i}\boldsymbol{\varepsilon}_{i}\left(\boldsymbol{\theta}\right)^{\top}\boldsymbol{G}_{i}^{-1}\right),
\]
\[
\nabla_{\boldsymbol{\alpha}}\rho_{i}\left(\boldsymbol{\theta},\mathcal{D}_{i}\right)=
\]
\[
\frac{1}{2}\left(\boldsymbol{z}_{i}^{\top}\left(\frac{\varepsilon_{1,i}\left(\boldsymbol{\theta}\right)^{2}}{g_{1,i}}-1\right),\;\boldsymbol{z}_{i}^{\top}\left(\frac{\varepsilon_{2,i}\left(\boldsymbol{\theta}\right)^{2}}{g_{2,i}}-1\right)\right)^{\top},
\]

\[
\boldsymbol{R}=\left(\begin{array}{cc}
\boldsymbol{I}_{2,-1} & \boldsymbol{O}_{2,2}\\
\boldsymbol{O}_{2,2} & \boldsymbol{I}_{2,-2}
\end{array}\right),
\]
where $\boldsymbol{I}_{a,-b}$ denotes the $a$-dimensional identity
matrix with the $b$th column being deleted and $\boldsymbol{O}_{a,b}$
denotes the $a\times b$ matrix of zeros.

\bibliographystyle{ieeetr}
\bibliography{reference}

\end{document}